\documentclass[conference]{IEEEtran}
\IEEEoverridecommandlockouts

\usepackage{amsmath,amsfonts,bm, physics}

\def\eqref#1{Eq.~(\ref{#1})}

\def\1{\bm{1}}

\def\mI{{\bm{I}}}

\DeclareMathAlphabet{\mathsfit}{\encodingdefault}{\sfdefault}{m}{sl}
\SetMathAlphabet{\mathsfit}{bold}{\encodingdefault}{\sfdefault}{bx}{n}

\def\*#1{\mathbf{#1}}
\def\$#1{\mathcal{#1}}
\def\^#1{\mathbb{#1}}

\newcommand{\R}{\mathbb{R}}

\usepackage[utf8]{inputenc} 
\usepackage[T1]{fontenc}    
\usepackage{hyperref}       
\usepackage{url}            
\usepackage{booktabs}       
\usepackage{amsfonts}       
\usepackage{nicefrac}       
\usepackage{microtype}      
\usepackage{lipsum}
\usepackage{fancyhdr}       
\usepackage{graphicx}       
\usepackage{caption}

\usepackage{xcolor}         
\usepackage{multirow}
\usepackage{tabularx}
\usepackage{subfigure}
\usepackage{bm}

\usepackage{amsmath,amssymb,amsthm}
\usepackage{algorithm}
\usepackage[noend]{algpseudocode}
\usepackage{cleveref}
\usepackage{balance}

\definecolor{light-gray}{gray}{0.92}

\graphicspath{{./figure_folder/}}

\def\BibTeX{{\rm B\kern-.05em{\sc i\kern-.025em b}\kern-.08em
    T\kern-.1667em\lower.7ex\hbox{E}\kern-.125emX}}

\begin{document}

\title{Textual and Visual Prompt Fusion for Image Editing via Step-Wise Alignment
}

\author{
  \IEEEauthorblockN{
    Zhanbo Feng\IEEEauthorrefmark{1}, 
    Zenan Ling\IEEEauthorrefmark{2}, 
    Xinyu Lu\IEEEauthorrefmark{1},
    Ci Gong\IEEEauthorrefmark{2}, 
    Feng Zhou\IEEEauthorrefmark{3},
    Wugedele Bao\IEEEauthorrefmark{4},\\
    Jie Li\IEEEauthorrefmark{1},
    Fan Yang\IEEEauthorrefmark{1},
    and Robert C. Qiu\IEEEauthorrefmark{2}
  } 
  \IEEEauthorblockA{\IEEEauthorrefmark{1}Department of Computer Science and Engineering, Shanghai Jiao Tong University} 
  \IEEEauthorblockA{\IEEEauthorrefmark{2}EIC,  Huazhong University of Science and Technology  \IEEEauthorrefmark{4}School of Computer Science, Hohhot Minzu College\\}

\thanks{
Z. Ling is supported by the National Natural Science Foundation of China (via NSFC-62406119), the Natural Science Foundation of Hubei Province (2024AFB074), and the Guangdong Provincial Key Laboratory of Mathematical Foundations for Artificial Intelligence (2023B1212010001).}

\thanks{
J. Li has been partially supported by NSFC Grants 61932014, 62232011 and the Shanghai Science and Technology Innovation Action Plan Grant 24BC3201200.}

\thanks{
W. Bao is partially Supported by Natural Science Foundation of Inner Mongolia Autonomous Region of China Grant 2021MS06001.}

\thanks{
R. C. Qiu would like to acknowledge the National Natural Science Foundation of China (via NSFC12141107), the Interdisciplinary Research Program of HUST (2023JCYJ012), the Key Research and Development Program of Guangxi (GuiKe-AB21196034). F. Zhou was supported by the National Natural Science Foundation of China Project (via NSFC-62106121) and the MOE Project of Key Research Institute of Humanities and Social Sciences (22JJD110001).
\textit{(Corresponding authors: \href{lingzenan@hust.edu.cn}{Zenan Ling} , \href{lijiecs@sjtu.edu.cn}{Jie Li})}}
}

\maketitle

\begin{abstract}
The use of denoising diffusion models is becoming increasingly popular in the field of image editing. However,  current approaches often rely on either image-guided methods, which provide a visual reference  but lack control over semantic consistency, or text-guided methods, which ensure alignment with the text guidance but compromise visual quality. To resolve this issue, we propose a framework that integrates a fusion of generated visual references and text guidance into the semantic latent space of a \textit{frozen} pre-trained diffusion model. Using only a tiny neural network, our framework provides control over diverse content and attributes, driven intuitively by the simple prompt. Compared to state-of-the-art methods, the framework generates images of higher quality while providing realistic editing effects across various benchmark datasets. The code is available at \url{https://github.com/SadAngelF/Editing-via-Step-Wise-Alignment}.
\end{abstract}


\section{Introduction}
\label{sec:intro}

Manipulating real-world images with natural language has long been a challenge in image processing.
Recently, denoising diffusion models (DDMs) have shown substantial success in text-to-image tasks, exemplified by models like Imagen~\cite{saharia2022photorealistic}, Dall-E~\cite{ramesh2021zero}, and Stable Diffusion~\cite{rombach2022high}. These text-to-image models produce diverse,  highly coherent, and realistic images that align well with text prompts. However, manipulating attributes on
real images is still a significant challenging.

With the advancement of Natural Language Processing (NLP) techniques, e.g., GPT~\cite{openai2023gpt4,brown2020language, google2023plam}, considerable effort has been invested in text-guided image editing~\cite{wallace2022edict,daras2022multiresolution}. 
Many previous works~\cite{bao2023one,hertz2022prompt,kim2022diffusionclip,wang2022imagen} have  developed image-editing techniques guided by textual prompts. However, they tend to neglect the importance of visual references.
Despite maintaining semantic fidelity, text-guided methods struggle to learn fine-grained visual patterns from textual features in the absence of a visual prior.
Textual semantics alone provide insufficient visual reference, leading to imprecise semantic manipulation. 
Text-guided editing is especially prone to failure when the target semantic is outside the domain.

On the other hand, image-guided editing can easily perform style transfer~\cite{choi2021ilvr, brooks2022instructpix2pix, nguyen2023visual,meng2021sdedit}, inpainting~\cite{lugmayr2022repaint}, and item replacement~\cite{jia2023taming,nitzan2022mystyle}. With visual reference, generators insert ready-made visual patterns into images directly.
Specifically, Taming Encoder~\cite{jia2023taming} embeds specified elements into the target image by encoding a reference image.
VISII~\cite{nguyen2023visual} blends both textual and visual prompts to learn a style transfer from paired examples, representing the “before” and “after” images of an edit. However, image-guided approaches lack intuitive control over semantic consistency, making it ambiguous to specify which attribute should be referenced from the image.

In this paper, we propose Step-Wise Alignment~(SWA) that integrates both visual references and text guidance into the semantic latent space of a \textit{frozen} pre-trained diffusion model. Our method leverages text guidance to provide intuitive control over semantic consistency, while refining the alignment between the text features and the semantic latent space of the diffusion model by incorporating a visual reference. Our main contributions are as follows:

\begin{enumerate}
  \item We introduce a framework that integrates a fusion of visual and textual prompts for attribute editing on real images.
  \item We propose Step-Wise Alignment to align the text-image fusion features and the semantic latent space of the \textit{frozen} diffusion model. Benefiting from zero-shot optimization, SWA avoids collecting the data of specific attributes.
  \item SWA is evaluated on various benchmark datasets, including CelebA-HQ, LSUN-church, and LSUN-bedroom, and it outperforms state-of-the-art methods in terms of image quality and attribute manipulation.
\end{enumerate}

\section{Methodology}

\subsection{Problem Definition}

Given an image $i_\text{edit}\in \R^{m\times n}$ and an attribute $t_\text{attr}$, our primary objective is to modify $i_\text{edit}$ according to the attribute $t_\text{attr}$,
resulting in an edited image, denoted as $i_\text{out}$.

Directly adding noise to $i_\text{edit}$ and performing denoising within a \textit{frozen} diffusion model is not a feasible approach for attribute editing.
The reason is that such a \textit{frozen} model may lack semantic relevance
and may fail to retain the desired attributes. In addition, the added noise can distort the image and introduce undesirable artifacts. Therefore, we propose a framework to refine the alignment between text features and the semantic latent space of the diffusion model by incorporating a visual reference.

Therefore, we aim to optimize the reverse process of a \textit{frozen} diffusion model for meaningful attribute editing. A typical reverse process in diffusion model is
\begin{eqnarray}
\label{eq_ddim}
    x_{t-1} = \sqrt{\alpha_{t-1}/ \alpha_t} \left(x_t - \sqrt{1-\alpha_t}\epsilon_\theta(x_t,t)\right) \nonumber \\  
     ~ + \sqrt{1-\alpha_{t-1}-\sigma_t^2} \cdot \epsilon_\theta (x_t,t) + \sigma_t \epsilon_t, 
\end{eqnarray}
where $\epsilon_t \sim \mathcal{N} (\mathbf{0},\mI)$ is standard Gaussian noise, $\alpha_t$ is the parameter based on the forward process, $\sigma_t = \eta \sqrt{(1-\alpha_{t-1})/(1-\alpha_t)}\sqrt{1-\alpha_t/\alpha_{t-1}}$, and $\epsilon_\theta (x_t,t)$ is a neural network to predict the noise in $x_t$. In the following, we introduce our framework to optimize the process above.

\subsection{Framework}

\begin{figure}
  \centering
  \begin{center}  
    \includegraphics[width=\columnwidth]{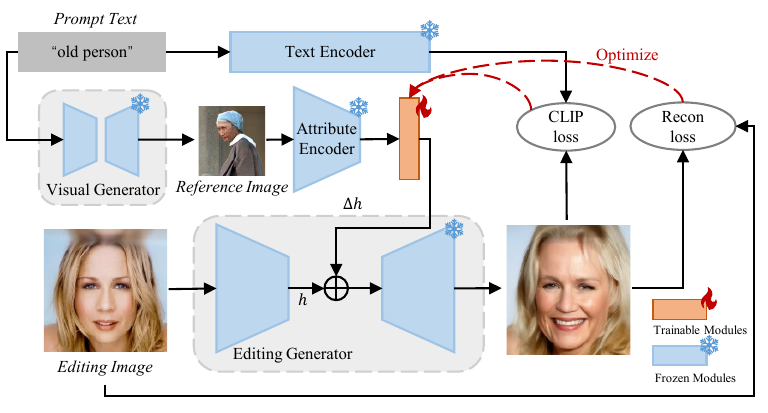}  
    \caption{\textbf{The framework of SWA.} 
     The reference image is encoded into features $\Delta h$. Then, $\Delta h$ are integrated into the latent features $h$ of the editing image. The textual prompt contributes semantic information for the manipulation process.}
    \label{fig_architecture}
  \end{center}  
\end{figure}

As illustrated in~\Cref{fig_architecture}, our framework consists of four key components: Text Encoder, Visual Generator, Attribute Encoder, and Editing Generator.
To obtain the visual attributes corresponding to the designated attribute, we utilize a text-image model as the Visual Generator. Furthermore, when the attribute involves adding embellishments, such as the glasses, a reference image can be manually provided to ensure consistency of embellishments during editing.
Both the textual prompt and the visual prompt, which are encoded by the Text Encoder and the Attribute Encoder respectively, are employed for editing in the Editing Generator.

\textbf{Text Encoder:} The text prompt $t$ from the attribute $t_\text{attr}$ is encoded into a vector for the purpose of calculating loss with the target image. 
In this paper, CLIP~\cite{radford2021learning} is used as the Text Encoder. 


In Text Encoder, let $E_T$ be a text encoder with vocabulary $V$. The attribute $t_\text{attr}$ is a sequence of phrases $t_\text{attr} = (s_1,\dots,s_k)$ with $s_i \in V$; for example, $k = 1$ if the attribute $t_\text{attr}$ is ``glasses.'' Similar to the prompt in NLP~\cite{schick2020exploiting}, we define a \textit{pattern} as a function $P$ that takes $t_\text{attr}$ as input and outputs two phrases or sentences $t_\text{source}, t_\text{target} = P(t_\text{attr})\in V$ as the text prompt $t = (t_\text{source}, t_\text{target})$. For example, the pattern $P(t_\text{attr})=$ \colorbox{light-gray}{(``a person'', ``a person with $t_\text{attr}$'')} will be used for the attribute $t_\text{attr}$ of the person. Given an  input attribute
$ t_\text{attr} = \colorbox{light-gray}{``glasses''}$,
then the text prompt will be
$  P(t_\text{attr})= \colorbox{light-gray}{(``a person'', ``a person with glasses'')}$.
After that, $t_\text{source} = \colorbox{light-gray}{``a person''}$ and $t_\text{target} = \colorbox{light-gray}{``a person with glasses''}$. Both of them compose the text prompt $t$. Using the Text Encoder, the text prompt will be encoded as $E_T(t_\text{source}) \in \R^d$ and $E_T(t_\text{target}) \in \R^d$.

\textbf{Visual Generator:} To obtain visual features of the designated attribute, a text-image model is used as the Visual Generator. Large generative models are known for strong robustness and generalization in conditional generation~\cite{saharia2022photorealistic, rombach2022high}.  
Despite their limitations in accurately detecting or modifying attributes, they can effectively generate the corresponding visual features.
In this paper, we use UniDiffuser~\cite{bao2023one} as the Visual Generator, which generates images by one model, benefiting from the marginal, conditional, and joint distributions determined by multi-modal data.
The reference image $i_\text{ref}$ is sampled from the conditional distribution $p(x_0|t_\text{target})$ and denoised by the noise predictor $\epsilon_\theta$ \cite{bao2023one}.

\textbf{Attribute Encoder:} The Attribute Encoder $E_A$ is a down-sampling network that incorporates attention blocks and residual blocks.
This type of down-sampling network is commonly used in both detection~\cite{ronneberger2015u} and generation~\cite{ho2020denoising} tasks. It encodes the original image into a latent space through down-sampling. The latent embedding of visual features is denoted as $E_A(i_\text{ref}) \in \R^D$, which is obtained
by taking the reference image as input.

\textbf{Editing Generator:} The Editing Generator generates the target image by inserting the fusion features $\Delta h$ into the latent space of a \textit{frozen} diffusion model. Previous studies have demonstrated the remarkable performance of diffusion models~\cite{ho2020denoising,choi2021ilvr} in this context. In this paper, we use DDIM~\cite{song2020denoising} to train a \textit{frozen} diffusion model as Editing Generator.

\subsection{Step-Wise Alignment}
\label{Sec:SWA}

In this section, we propose Step-Wise Alignment (SWA) to align the fusion prompt from the Attribute Encoder and Text Encoder. SWA optimizes the Attribute Encoder to align the latent embedding of the editing image with the reference image, guided by the embedding of the textual prompt.

A straightforward approach to perform latent manipulation during the generation of $x_0$ from $x_T$ is to update the Attribute Encoder to minimize the following loss:
\begin{equation}
    \label{eq_l_direction}
    \mathcal{L}_\text{dir} (i_\text{out}, t_\text{target}; i_\text{edit}, t_\text{source}) := 1- \frac{\Delta I \cdot \Delta T}{\| \Delta I \|  \| \Delta T \|}, 
  \end{equation}
where $\Delta I = E_A(i_\text{out})- E_A(i_\text{edit})$ and $\Delta T = E_T(t_\text{target})- E_T(t_\text{source})$, for the generated image $i_\text{out}$,  the editing image $i_\text{edit}$, the target prompt $t_\text{target}$, and the source prompt  $t_\text{source}$.

However, this approach might cause image distortion or
erroneous manipulations, as observed in prior works~\cite{choi2021ilvr,mokady2022null}.
An alternative approach entails adjusting the noise $\epsilon_t^\theta$ predicted by the network during each sampling iteration. In brief, we can reformulate the diffusion process of DDIM as follows:
\begin{equation}
  \label{eq_ddim_short}
  x_{t-1} = \sqrt{\alpha_{t-1}} P_t(\epsilon_\theta(x_t,t))  + D_t(\epsilon_\theta(x_t,t)) + \sigma_t \epsilon_t,
\end{equation}
where $P_t(\epsilon_\theta(x_t,t))= \frac{1}{\sqrt{\alpha_t}}\left(x_t - \sqrt{1-\alpha_t}\epsilon_\theta(x_t,t) \right)$ as the predicted $x_0$, and $D_t(\epsilon_\theta(x_t,t)) = \sqrt{1-\alpha_{t-1}-\sigma_t^2} \cdot \epsilon_\theta (x_t,t)$ as the direction to $x_t$.
Nonetheless, making direct modifications to the noise $\epsilon_\theta$ in both $P_t$ and $D_t$ leads to mutual nullification, yielding an unchanged $p_\theta(x_{0:T})$. This phenomenon mirrors a form of destructive interference, as elucidated in Asyrp~\cite[Theorem~1]{kwon2022diffusion}, which will inadvertently nullify the effects of optimizing $\epsilon_\theta$.

\begin{algorithm}[t]
\caption{Image Editing via SWA}
\label{alg:ZIP}
\begin{algorithmic}[1]
\Require 
    \begin{tabular}[t]{@{}l@{}}
    \textbf{An editing image} $i_\text{edit}$; \textbf{A text prompt} $t_\text{attr}$; \textbf{Editing}\\ \textbf{Generator}; $\epsilon_\theta$;\textbf{Visual Generator} $G_V$; \textbf{Attribute}\\ \textbf{Encoder} $E_A$; \textbf{CLIP encoder} $\xi_{clip}$; \textbf{Diffusion model}\\ \textbf{timestep} $T$; \textbf{Timestep} $t_\text{swa}$
    \end{tabular}  
\Ensure
    \begin{tabular}[t]{@{}l@{}}
    A target image $i_\text{out}$
    \end{tabular}  
\State Initialize $t_\text{source}$ and $t_\text{target}$ based on $t_\text{attr}$. 
\State Generate the reference image $i_\text{ref}=G_V(t_\text{target})$. 
\State Encode $\Delta h = E_A(i_\text{ref})$ .                           
\State Get the noise image $x_0$ from $i_\text{edit}$ based on $\epsilon_\theta$.
\For{$i=1,2,\ldots,N$}
    \For{$t=T,T-1\ldots,0$}
    \hspace{-5pt}\If {$t > t_\text{swa}$}
        \\ \hspace{23pt}  $x_{t-1} = \sqrt{\alpha_{t-1}} P_t(\widetilde{\epsilon}_\theta(x_t,t))  + D_t(\epsilon_\theta(x_t,t)) + \sigma_t \epsilon_t.$ 
    \hspace{-5pt}\Else
        \\ \hspace{23pt}  $x_{t-1} = \sqrt{\alpha_{t-1}} P_t(\epsilon_\theta(x_t,t))  + D_t(\epsilon_\theta(x_t,t)) + \sigma_t \epsilon_t.$ 
    \EndIf
    \EndFor
    \State  $i_\text{out} \xleftarrow{} x_0$ .
    \State Update the parameters of Attribute Encoder $E_A$ as~\Cref{eq_loss}.
\EndFor
\State \Return $i_\text{out}$
\end{algorithmic}

\end{algorithm}

Hence, in order to circumvent the interference delineated in~\Cref{eq_ddim_short}, we adopt an asymmetrical form in SWA:
\begin{equation}
  \label{eq_zip}
  x_{t-1} = \sqrt{\alpha_{t-1}} P_t(\widetilde{\epsilon}_\theta(x_t,t))  + D_t(\epsilon_\theta(x_t,t)) + \sigma_t \epsilon_t.
\end{equation}
Here, $\widetilde{\epsilon}_\theta(x_t,t)$ represents an adjustment to
$\epsilon_\theta(x_t,t)$ grounded on the visual features $\Delta h$. This is achieved by introducing $\Delta h$ into the original feature maps $h_t$ derived from $x_t$.

The optimization of $\epsilon_\theta$ to $\widetilde{\epsilon}_\theta$ is achieved by using the text prompt $t$ to guide the generation process.
Due to the absence of ground truth for editing images in editing tasks, fully supervised training methods are not applicable. Given that CLIP has the capability for vision-language alignment and can effectively evaluate the editing results without ground-truth images, we use CLIP loss to fine-tune the Attribute Encoder.

Following the approach presented in~\cite{avrahami2022blended}, we employ the directional CLIP loss in~\Cref{eq_l_direction} as our loss function:
\begin{equation}
  \label{eq_loss}
  \mathcal{L} = \lambda_\text{clip} \mathcal{L}_\text{dir} (\widetilde{P_t}, t_\text{target}; P_t, t_\text{source}) +  \lambda_\text{recon} | x_\text{out}^t - x_\text{edit}^t |.  
\end{equation}
The modified $\widetilde{P_t}$ and the original $P_t$ correspond to the formulations presented in~\Cref{eq_zip} and~\Cref{eq_ddim_short}, respectively.
The last term in \Cref{eq_loss} is the reconstruction loss, 
calculated as an
$\ell_1$ loss between the generated image and the original image.
It effectively preserves the original features, preventing drastic alterations. To balance the aforementioned losses, we introduce the hyperparameters $\lambda_\text{clip}$ and $\lambda_\text{recon}$.

\subsection{Image Editing via SWA}
\label{sec:image_editing_zip}

Given an image $i_\text{edit}\in \R^{m\times n}$ and an attribute $t_\text{attr}$, the visual features from the reference image are integrated into the latent space of $i_\text{edit}$ as detailed in~\Cref{eq_zip}, and the textual prompt is used to optimize the Attribute Encoder as~\Cref{eq_loss}. Meanwhile, the diffusion model (Editing Generator) remains \textit{frozen}. The whole process is shown in Algorithm~\ref{alg:ZIP}.

\section{Experiments}
\label{sec_exp}


\noindent\textbf{Evaluation.}
Various metrics have been introduced in prior research to evaluate the effectiveness of image generation and editing.
In this study, we employ the Inception Score (ISC)~\cite{salimans2016improved} and the Fréchet Inception Distance (FID)~\cite{szegedy2016rethinking} as indicators of the image generation quality. Furthermore, we leverage the CLIP Score~\cite{radford2021learning} to assess
the alignment between edited images and their intended semantic targets.

\begin{figure}[t]
  \centering
  \includegraphics[width=0.5\columnwidth]{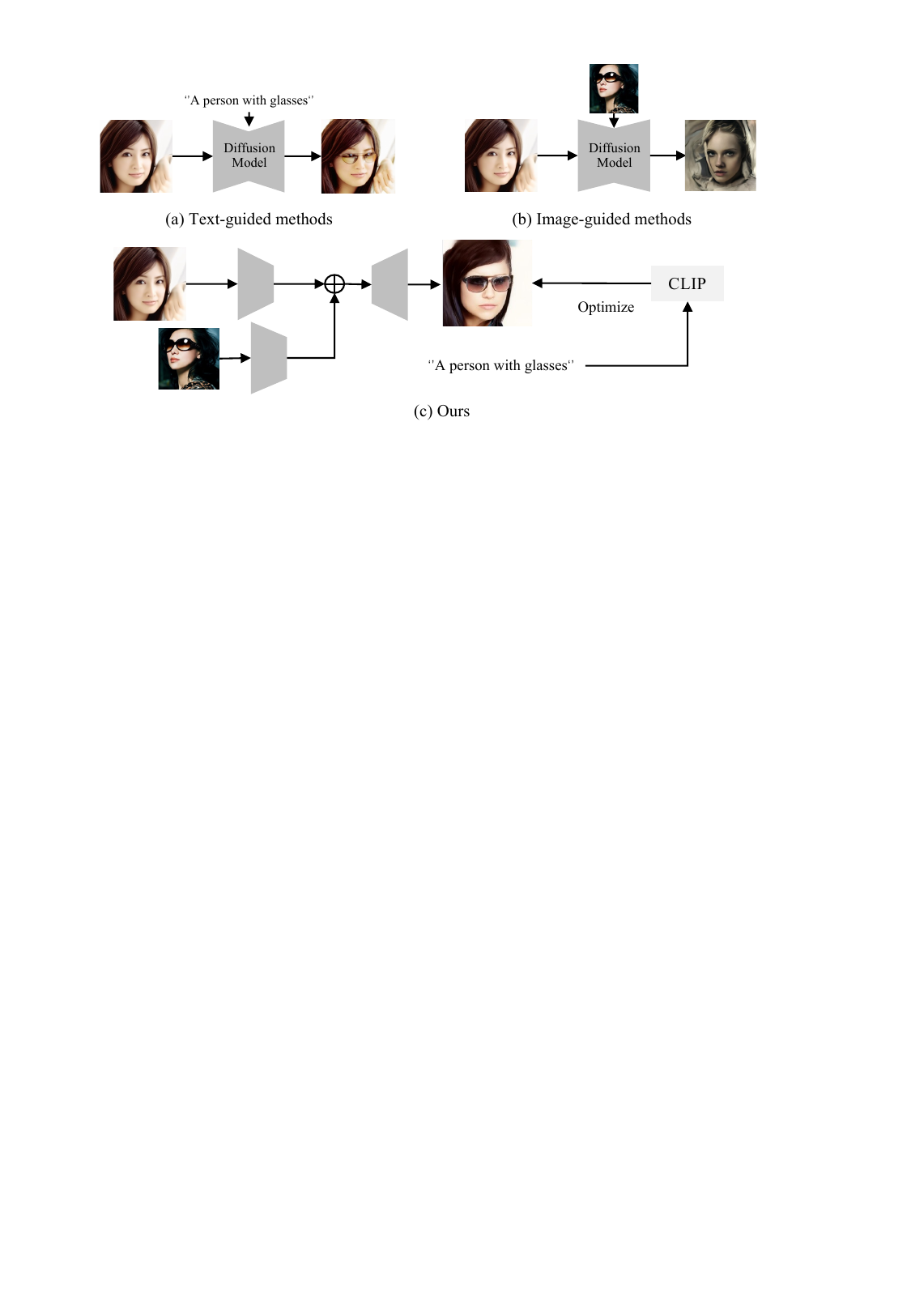}
  \caption{\textbf{Consistent editing:} 
  Once a reference image is provided, our method enables
  consistent and controllable editing to the reference image. 
  In contrast, NTI fails to
  generate the corresponding style for the glasses.}
  \vspace{-15pt}
  \label{fig:consistent}
\end{figure}

\begin{figure}
  \centering
  \includegraphics[width=0.7\columnwidth]{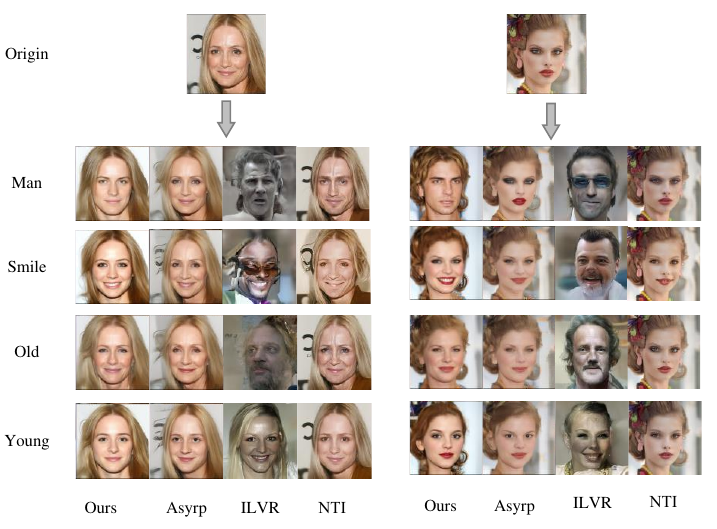}
  \caption{\textbf{Editing results for in-domain attributes}}
  \vspace{-15pt}
  \label{fig_indomain}
\end{figure}

\subsection{Editing Consistency}
\label{sec:consistence}

Our framework ensures high consistency between the attributes of editing images and their reference images. Although previous works, such as NTI~\cite{mokady2022null}, achieve realistic edits for real images, controlling the style of attributes remains challenging. As shown in~\Cref{fig:consistent}, 
SWA generates consistent styles of glasses across different images when given a specific reference image, whereas Null-Text Inversion cannot provide this level of control.

\subsection{Editing Generalization}
\label{sec_indomain}

\begin{figure}
    \centering 
    \subfigure[Makeup]
    {  \includegraphics[width=0.35\columnwidth]{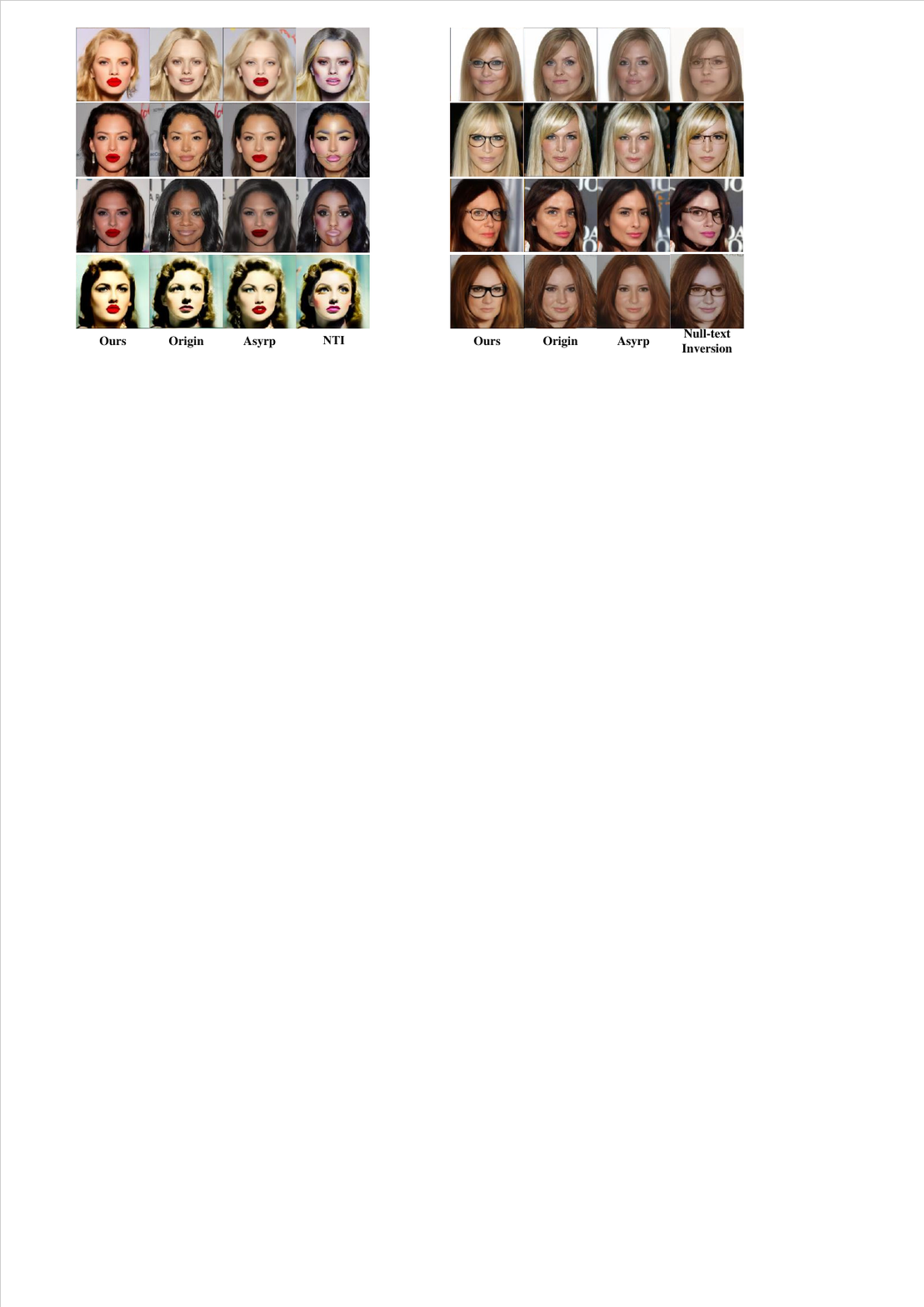}  }  
    \subfigure[Glasses]
    {  \includegraphics[width=0.35\columnwidth]{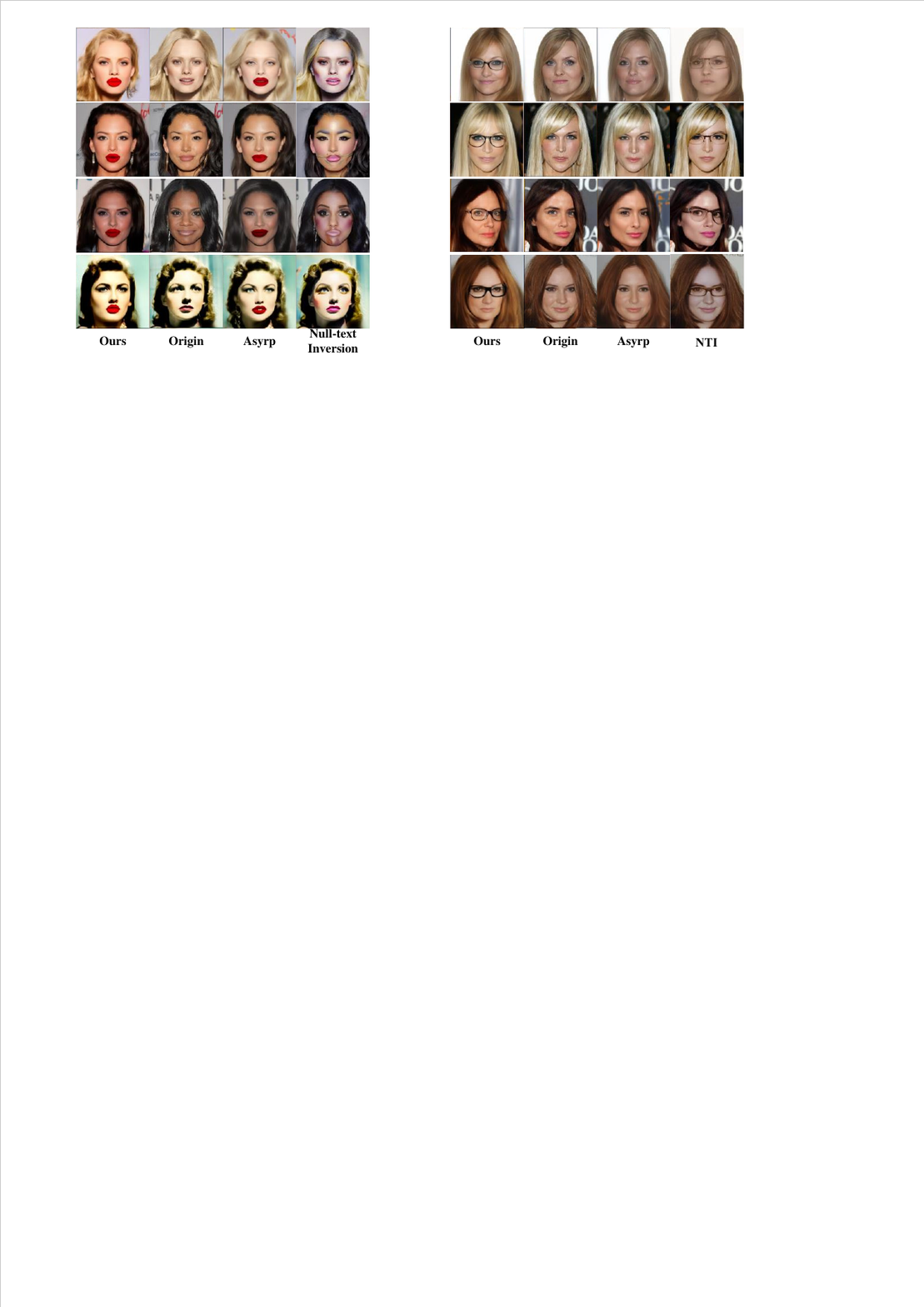}  }  
    \caption{\textbf{Editing results for out-of-domain attributes.}}
    \label{fig_outdomain}
\end{figure}

\begin{figure}[t]
  \centering
\includegraphics[width=0.75\columnwidth]{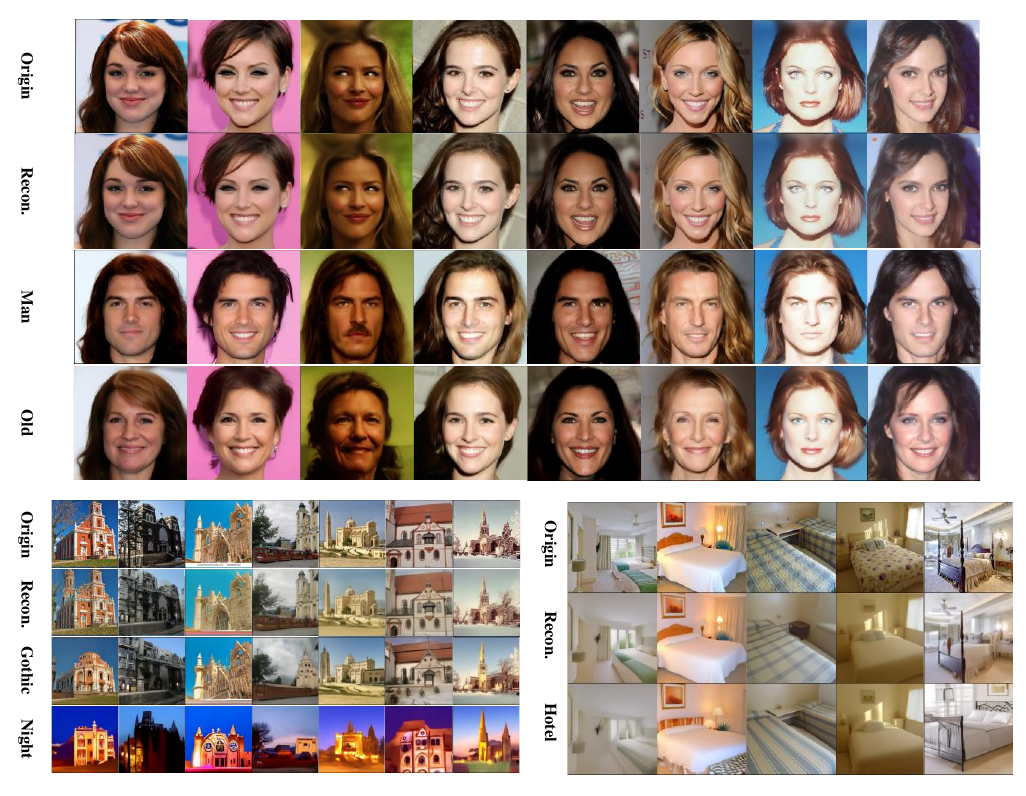}
  \caption{\textbf{Editing results of SWA on various datasets.}}
  \vspace{-15pt}
  \label{fig_allexp}
\end{figure}

Both in-domain and out-of-domain attributes can be edited using our method. In-domain attributes refer to features that the \textit{frozen} diffusion model has encountered during training.
For instance, in the CelebA-HQ dataset, 
many images depict individuals with smiling expressions,
and the attribute ``smiling'' is explicitly labelled in the dataset.
On the other hand, out-of-domain attributes, such as ``Add glasses'', are not represented in the training data.
As shown in \Cref{fig_indomain} and \Cref{tab_indomain}, SWA demonstrates strong performance for in-domain attributes. Similarly, \Cref{fig_outdomain} and \Cref{table_outdomain1} showcase SWA's ability to achieve high quality editing results for out-of-domain attributes as well.
\Cref{fig_allexp} illustrates the visual results attained through SWA across various datasets.

\begin{table}
  \centering
  \caption{In-domain Attributes Modification}
  \label{tab_indomain}
  \resizebox{\columnwidth}{!}{
    \begin{tabular}{cccccc}
    \toprule
                             &         & ILVR~\cite{choi2021ilvr} & Asyrp~\cite{kwon2022diffusion} & NTI~\cite{mokady2022null} & Ours \\ \midrule
    \multirow{3}{*}{Man}     &  ISC \textuparrow   & \underline{2.623} & 1.808 & 2.219 & \textbf{3.292} \\
                             &  FID \textdownarrow & 150.2             & 77.65 & \underline{52.23} & \textbf{45.31} \\
                             &  CLIP \textuparrow  & 22.24             & 19.15 & \textbf{23.02} & \underline{22.38} \\ 
                             \midrule
    \multirow{3}{*}{Old}     &  ISC \textuparrow   & 2.361             & 1.833 & 2.075 & \textbf{2.778} \\ 
                             &  FID \textdownarrow & 145.1             & 77.82 & \textbf{50.65} & \underline{55.96} \\
                             &  CLIP \textuparrow  & \textbf{23.26}    & 22.73 & 22.46 & \underline{22.79} \\ 
                             \midrule
    \multirow{3}{*}{Smiling} &  ISC \textuparrow   & \underline{2.557} & 1.760 & 2.014 & \textbf{2.573} \\ 
                             &  FID \textdownarrow & 223.7             & \underline{73.93} & \textbf{44.25} & 85.36 \\
                             &  CLIP \textuparrow  & 25.59             & \underline{26.31} & 25.98 & \textbf{27.01} \\ \midrule
    \multirow{3}{*}{Young}   &  ISC \textuparrow   & \underline{2.582} & 1.705 & 2.089 & \textbf{2.624} \\ 
                             &  FID \textdownarrow & 147.9             & 80.03 & \textbf{40.52} & \underline{55.28} \\
                             &  CLIP \textuparrow  & 22.70             & \underline{26.11} & 24.42 & \textbf{25.07} \\ \bottomrule
    \end{tabular}
  }
\end{table}

\begin{figure}
  \centering
    \includegraphics[width=0.8\columnwidth]{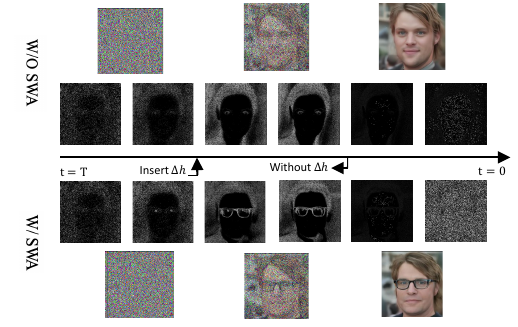}
    \caption{\textbf{Ablation experiments of reference image: } The top half depicts the process without a reference image, while the bottom includes a reference image. Pixels are more $concentrated$ on the attribute's features.}
    \vspace{-10pt}
  \label{fig_visual}
\end{figure}

\subsection{Ablation Experiments}

We depict the attribute editing process for the attribute ``glasses'' facilitated by SWA.
\Cref{fig_visual} effectively demonstrates that pixels are more \textit{concentrated} on attribute features when a reference image is used (with SWA), highlighting the effectiveness of incorporating a reference image in generating visual features for the target attribute.

\begin{table}[h!]
  \begin{center}
  \caption{Out-of-domain Attributes Modification}
  \label{table_outdomain1}
  \resizebox{\columnwidth}{!}{
  
    \begin{tabular}{ccccccc}
    \toprule
    
                             &         & ILVR~\cite{choi2021ilvr} & Asyrp~\cite{kwon2022diffusion} & NTI~\cite{mokady2022null} & Ours \\ \midrule
    \multirow{3}{*}{Makeup}  &  ISC \textuparrow     & 1.940 & 1.755 & \underline{1.976} & \textbf{2.467} \\
                             &  FID \textdownarrow   & 144.4 & \underline{77.49} & \textbf{69.64} & 88.0 \\
                             &  CLIP \textuparrow    & 24.10 & 23.86 & \textbf{26.46} & \underline{25.12} \\ \midrule
    \multirow{3}{*}{Glasses} &  ISC \textuparrow     & \textbf{3.060} & 1.390 & 1.980 & \underline{2.418} \\ 
                             &  FID \textdownarrow   & 165.6 & 147.5 & \textbf{47.55} & \underline{124.6} \\
                             &  CLIP \textuparrow    & 24.52 & \underline{29.55} & 26.17 & \textbf{30.07} \\ \bottomrule
    \end{tabular}
    }
    \end{center}
    \vspace{-10pt}
\end{table}

\section{Conclusion}

This paper introduces SWA, a novel approach for manipulating real-world images by fusing textual and visual prompts. SWA integrates a blend of generated visual reference and textual guidance into the semantic latent space of a \textit{frozen} diffusion model. By bridging the gap between visual patterns and textual semantics, SWA effectively alters both in-domain and out-of-domain attributes. 
In future work, our research will focus on improving the precision of attribute extraction from reference images.
Specifically, we aim to refine methods for distinguishing attributes with similar visual features, thereby further improving the manipulation process.

\balance
\bibliographystyle{IEEEtran}
\bibliography{fzb}

\end{document}